\documentclass{article}

\usepackage{arxiv}

\usepackage{hyperref}       
\usepackage{url}            
\usepackage{amsfonts}       
\usepackage[round]{natbib}
\usepackage{graphicx}
\usepackage{xcolor}
\usepackage{mathtools, amsthm}
\usepackage{amsmath}
\usepackage{amssymb}
\usepackage{MnSymbol}
\usepackage{tikz}
\usepackage{enumitem}

\usepackage[]{color}
\usepackage{alltt}
\usepackage{algorithm}
\usepackage{algpseudocode}
\usepackage{array}
\usepackage{bm}
\usepackage{booktabs}
\usepackage{amssymb,amsmath,amsthm}
\usepackage[utf8]{inputenc} 
\usepackage[T1]{fontenc}    
\usepackage{hyperref}       
\usepackage{url}            
\usepackage{booktabs}       
\usepackage{amsfonts}       
\usepackage{nicefrac}       
\usepackage{microtype}      
\usepackage{xcolor}         
\usepackage{mathtools}
\usepackage{mathbbol}

\numberwithin{equation}{section}
\usepackage{parskip}
 
\DeclareMathOperator*{\argmax}{arg\,max} 
\DeclareMathOperator{\E}{\mathbb{E}}
\usepackage{natbib}

\usepackage{ifthen}
\usepackage{graphicx}

\newcommand{\compilehidecomments}{false} 
\ifthenelse{ \equal{\compilehidecomments}{true} }{%
	\newcommand{\yiliu}[1]{}
	\newcommand{\wei}[1]{}
	\newcommand{\milan}[1]{}
}{
	\newcommand{\yiliu}[1]{{\color{red}  [#1]}}
	\newcommand{\wei}[1]{{\color{blue} [#1]}}
	\newcommand{\milan}[1]{{\color{cyan} [#1]}}
}

\newcommand{\kmax}{$k$-{\sc max}\ }
\newcommand{\arms}{\mathcal{A}}

\newcommand{\compilefullversion}{full} 
\ifthenelse{\equal{\compilefullversion}{false}}{%
	\newcommand{\OnlyInFull}[1]{}
	\newcommand{\OnlyInShort}[1]{#1}
}{
	\newcommand{\OnlyInFull}[1]{#1}%
	\newcommand{\OnlyInShort}[1]{}%
}

\setcitestyle{authoryear} 
\theoremstyle{plain}
\newtheorem{theorem}{Theorem}[section]

\newtheorem{lemma}[theorem]{Lemma}

\theoremstyle{definition}
\newtheorem{definition}[theorem]{Definition}

\theoremstyle{remark}

\renewcommand{\headeright}{}
\renewcommand{\undertitle}{}
\renewcommand{\shorttitle}{\textit{$k$-max bandits with max value-index feedback}}

\date{}

\title{Combinatorial Bandits for Maximum Value Reward Function  under Max Value-Index Feedback}

\author{Yiliu Wang\thanks{Department of Statistics, London School of Economics, London, United Kingdom, \url{y.wang298@lse.ac.uk}},  Wei Chen\thanks{Microsoft Research Asia, Beijing, China, \url{weic@microsoft.com}}, and Milan Vojnovi\' c\thanks{Department of Statistics, London School of Economics, London, United Kingdom, \url{m.vojnovic@lse.ac.uk}}
}

\begin{document}
	\maketitle
	
	\begin{abstract}
		We consider a combinatorial multi-armed bandit problem for maximum value reward function under maximum value and index feedback. This is a new feedback structure that lies in between commonly studied semi-bandit and full-bandit feedback structures. We propose an algorithm and provide a regret bound for problem instances with stochastic arm outcomes according to arbitrary distributions with finite supports. 
  The regret analysis rests on considering an extended set of arms, associated with values and probabilities of arm outcomes, and applying a smoothness condition. Our algorithm achieves a $O((k/\Delta)\log(T))$ distribution-dependent and a $\tilde{O}(\sqrt{T})$ distribution-independent regret where $k$ is the number of arms selected in each round, $\Delta$ is a distribution-dependent reward gap and $T$ is the horizon time. Perhaps surprisingly, the regret bound is comparable to previously-known bound under more informative semi-bandit feedback. We demonstrate the effectiveness of our algorithm through experimental results.
 
	\end{abstract}

	\section{Introduction}
	We consider a combinatorial bandit problem where in each round the agent selects a set of $k$ arms from a ground set of $n$ arms, with each selected arm yielding an independent random outcome, and the reward of the selected set of arms being the maximum value of these random outcomes.
	After selecting a set of arms in a round, the agent observes the feedback that consists of the maximum outcome value and the identity of the arm that achieved this value.
We refer to this sequential decision making problem as \kmax \emph{bandit with max value-index feedback}.
	The outcomes of arms are assumed to be according to independent random variables over arms and rounds. We first consider arm outcomes according to binary distributions and then extend to arbitrary discrete distributions with finite supports.
The performance of the agent is measured by the expected cumulative regret over a time horizon, defined as the difference of the cumulative reward achieved by selecting a set with maximum expected reward in each round and the cumulative reward achieved by the agent. 


As a concrete motivating application scenario, consider an online platform recommending some products, e.g., Netflix, Amazon, and Spotify. Assume there is in total $n$ products for recommendation on the platform. 
The goal is to display a subset of $k$ product items to a user that best fit the user preference. The feedback can be limited, as we may only observe on which displayed product item the user clicked and their subsequent rating of this product. Many other problems can be formulated in our setting, such as project portfolio selection \citep{Blau2004, jekunen}, team formation \citep{KR18,SVY21,lee2021test,Mehta} and sensor placement problems \citep{Golovin2011, Asadpour2016}.

The goal of the agent is to maximize the expected cumulative reward over a time horizon. The problem is challenging mainly for two reasons. Firstly, the reward function is the maximum value function, which is nonlinear and thus depends not only on the expected values of the constituent base arms. The uncertainty of binary-valued arm outcomes makes the problem more challenging under the maximum value reward. As we will show in the numerical section, high-risk high-reward arms may outperform stable-outcome arms in this case. The second challenge is due to the limited feedback. The agent only observes the maximum value and the identity of an arm that achieves this value, which makes it difficult to learn about other non-winning arms. 

The problem we consider has actions and rewards as in some combinatorial bandits \cite{CesaBianchi2012, Chen2013, chen2016combinatorial}, but the max value-index feedback structure is neither the semi-bandit nor the full-bandit feedback, which are commonly studied in bandits literature. 
To elaborate on this, let $X_1, \ldots, X_n$ be independent random variables corresponding to arm outcomes. Then, for any set $S$ of arms, the semi-bandit feedback consists of all values $\{X_i\mid i\in S\}$, while the full-bandit feedback is only the maximum value, $\max\{X_i\mid i\in S\}$, under maximum value reward function. On the other hand, the max value-index feedback consists of the maximum value $\max\{X_i\mid i\in S\}$ and the index $I\in \argmax\{X_i\mid i\in S\}$. This feedback lies between semi-bandit feedback and full-bandit feedback, and is only slightly more informative than the full-bandit feedback. Indeed, the only information that can be deduced from the max value-index feedback about arms $j\neq I$, is that their outcome values are smaller than or equal to the observed maximum value $X_I$. The index feedback is alike to comparison observations in dueling bandits \citep{Ailon2014, Sui2017}, but with additional max value feedback.

We present algorithms and regrets bounds for the \kmax problem with max value-index feedback, under different assumptions on information available about arm outcomes. 
Our results show that despite considerably more limited feedback, comparable regret bounds to combinatorial semi-bandits can be achieved for the \kmax bandit problem with max value-index feedback.



\subsection{Related work}



The problem we study has connections with combinatorial multi-armed bandits (CMAB) \cite{CesaBianchi2012, Chen2013, chen2016combinatorial}. 
Most of the existing work on CMAB problems is focused on semi-bandit feedback setting, e.g. \cite{Chen2013, Kveton2015}. 
The \kmax problem with the semi-bandit feedback was studied in~\cite{Chen2016} whose solution is easier than in our paper because of the semi-bandit feedback assumption.

In most works on full-bandit CMAB problems, restrictions are placed on the reward function. 
\cite{Rejwan2020} considered 
the sum reward function. Only a few algorithms have been proposed for non-linear rewards. \cite{Katariya2017} considered minimum value reward function under assumption that arm outcomes are according to Bernoulli distributions, with analysis largely depending on this assumption.   
\cite{Gopalan2014} studied the full-bandit CMAB with general rewards using a Thompson sampling algorithm. However, it is computationally hard to compute the posteriors in the algorithm and the regret bound has a large exponential constant. Recent work by \cite{Agarwal2021} proposed a merge and sort algorithm 
under assumption that distributions of arm outcomes obey a first-order stochastic dominance (FSD) condition. This condition is restrictive, e.g. it fails to hold for binary distributions. We can conclude that full-bandit CMAB solutions proposed so far either do not apply to our problem or have exponential computational complexity.

A related work is on combinatorial cascading bandits, e.g. \cite{Kveton2015a}. 
Here the agent chooses an ordered sequence from the set of arms and the outcomes of arms are revealed one by one until a stopping criteria is met. 
\cite{chen2016combinatorial} generalized the problem to combinatorial semi-bandits with probabilistically triggered arms (CMAB-T). The main difference with our setting is that CMAB-T assumes more information and is inherently semi-bandit. By revealing the outcomes of arms one by one, the agent is able to observe outcomes of arms selected before the one meeting the criteria. Another difference is that we consider more general distributions of arm outcomes. 

We summarize some known results on regret bounds for CMAB problems in Table~\ref{tab:fprop}. In the table, $\Delta$ denotes the gap between the optimum expected regret of a set and the best suboptimal expected regret of a set. 
\cite{Katariya2017} considers a bipartite setting where the agent chooses a pair of arms from a row of $K$ items and a column of $L$ items. \cite{Agarwal2021} only provides a distribution-independent regret bound, which is worse than  $\tilde{O}(\sqrt{T})$ distribution-independent regret bounds that follow from our distribution-dependent regret bounds. 

Another related line of work is that on dueling bandits \cite{Ailon2014} where the agent plays two arms at each time and observes the outcome of the duel. The goal is to find the best arm in the sense of a Condorcet winner under relative feedback of the dueling outcomes. \cite{Sui2017} extended the setting to multiple dueling bandits problem by simultaneously playing $k$ arms instead of two arms. Compared with this line of work, we assume additional absolute feedback. Our goal is different as it requires selecting a best set of arms with respect to a non-linear reward function. 


Finally, our work is related to choice models, e.g. \cite{Luce59} \cite{Thurstone}, and sequential learning for choice models \cite{choicebandits}. The main difference with previous work is that we consider maximum value and index feedback. 





\begin{table*}[t]
	\begin{center}
		\caption{Known regret bounds for CMAB problems under different settings.}
		\label{tab:fprop}
		\begin{tabular}{c | c c c c}
			\hline
			&  Feedback & Reward & Assumptions & Regret \\
			\hline
			\cite{Chen2016}  &semi-bandit&general&general  &$O(\frac{nk}{\Delta}\log(T))$\\
			\cite{Rejwan2020}&full-bandit&linear&subgaussian  &$O(\frac{nk}{\Delta}\log(T))$\\
			\cite{Katariya2017}&full-bandit&min&Bernoulli&$O(\frac{K+L}{\Delta}\log(T))$\\
			\cite{Agarwal2021}&full-bandit&general&FSD&$\tilde{O}(n^{1/3}k^{1/2}T^{2/3})$\\
			\cite{Kveton2015a}&cascading-bandit&max&Bernoulli&$O(\frac{nk}{\Delta}\log(T))$\\
            Our work&value-index&max&general&$O(\frac{nk}{\Delta}\log(T))$\\
			\hline
		\end{tabular}\\
	\end{center}
\end{table*}

\subsection{Summary of contributions}

Our contributions can be summarized in the following points.

$\bullet$ We formulate and study a new combinatorial bandit problem for maximum value reward function under max value-index feedback. This feedback structure lies in between commonly studied full-bandit and semi-bandit feedback, and is only slightly stronger than full-bandit feedback. Compared to the full-bandit setting, we assume additional information of maximum-value index, which is observed in some real-world applications. 
Our work may be seen as a step towards solving full-bandit CMAB problems with non-linear reward functions under mild assumptions.



$\bullet$ We first present algorithms for binary distributions of arm outcomes and then use them to extend to the more general case of arbitrary discrete distributions with finite supports. For the case when the order of arm outcome values is a priori known, we show a Combinatorial Upper Confidence Bound (CUCB) algorithm and a regret bound that is comparable to that achievable under more informative semi-bandit feedback. 

$\bullet$ For the case when the ordering of values is a priori unknown to the algorithm, we show a variant of the CUCB algorithm defined by using the concept of arm equivalence. We show that this algorithm has a regret upper bound that is comparable to that shown to hold for the case of known ordering of values, and, thus, is comparable to the one achievable under semi-bandit feedback. 




\noindent
{\bf Organization of the paper. }
In Section \ref{sec:form}, we formally define the problem. In Section~\ref{sec:algo}, we first prove some key properties of reward function and then present our algorithms and regret bounds for the case of binary distributions of arm outcomes. In Section~\ref{sec:ext}, we discuss extension to arbitrary discrete distributions with finite supports. Section~\ref{sec:numeric} contains our numerical results. Finally, we summarize our work in Section~\ref{sec:conc}. Proofs of theorems are provided in Appendix.

\section{Problem formulation}\label{sec:form}


We consider a sequential decision making problem with an agent and a set of $n$ arms, denoted as $\arms = [n] = \{1,2,\ldots,n\}$. For each arm $i\in \arms$, outcomes are independent and identically distributed over rounds, according to a random variable $X_i$ with a discrete distribution with finite support. Let $0=v_{i,0} < v_{i,1} < \cdots < v_{i,s_i}$ denote values of the support of distribution of $X_i$, where $s_i$ is a positive integer, and $s_i+1$ is the support size. 
Let $p_{i,j} = \Pr[X_i = v_{i,j}]$ for $j \in \{0,1,\ldots, s_i\}$, with  $0 < \sum_{j=1}^{s_i} p_{i,j}\leq 1$.
Let $\bm{v} = (\bm{v}_1,\ldots, \bm{v}_n)$ and $\bm{p} = (\bm{p}_1,\ldots, \bm{p}_n)$ where $\bm{v}_i = (v_{i,1},\ldots, v_{i,s_i})$ with $v_{i,j}\in [0,1]$,
and $\bm{p}_i = (p_{i,1},\ldots, p_{i,s_i})$.  
For the special case of binary distributions, we write $p_{i}$ and $v_i$ in lieu of $p_{i,1}$ aƒnd $v_{i,1}$, respectively. 
Both $\bm{v}$ and $\bm{p}$, as well as the $s_i$'s in the general case, are unknown parameters to the agent. 



We define $\mathcal{F} = \{S\in 2^\arms\mid  |S| = k\}$ as the set of actions, where $0 < k \leq n$ is an integer. At each round $t$, the agent selects an action $S_t\in\mathcal{F}$. The agent observes the maximum outcome value of selected arms and the index of an arm achieving this maximum outcome value, and receives the reward corresponding to the maximum outcome value. 
We denote the expected reward of an action $S$ as $r_S(\bm{p},\bm{v}) = \E[\max\{X_i\mid  i\in S\}]$, which is a function of parameters $\bm{p}$ and $\bm{v}$. 
The performance of the agent is measured by the cumulative regret, defined as the difference of the expected cumulative reward achieved by playing the best action and the expected cumulative reward achieved by the agent. Denote $\mathrm{OPT}(\bm{p},\bm{v}) = \max\{r_S(\bm{p},\bm{v}) \mid S\in \mathcal{F}\}$. 
An $(\alpha,\beta)$-approximation oracle takes  $(\bm{p}',\bm{v}')$ as input and returns a set $S$ such that
$
\Pr[r_S(\bm{p}',\bm{v}')\geq \alpha\ \mathrm{OPT}(\bm{p}',\bm{v}') ]\geq \beta
$
where $\alpha$ is the approximation ratio and $\beta$ is the success probability. 
If the agent uses an $(\alpha,\beta)$-approximation oracle, then we consider the $(\alpha,\beta)$-approximation regret defined as
$$
R(T) = T\ \alpha\ \beta\ \mathrm{OPT}(\bm{p},\bm{v}) - \E\left[\sum_{t=1}^{T} r_{S_t}(\bm{p},\bm{v})\right].
$$


The offline $k$-MAX problem can be approximately solved by a greedy algorithm to achieve a $(1-1/e)$ approximate solution, or by a polynomial-time approximation scheme (PTAS)
to achieve a $(1-\varepsilon)$ approximate solution for any $\varepsilon > 0$~\cite{Chen2016}.
For the special case of binary distributions, an exact solution
can be found by using a dynamic programming algorithm \citep{YZ}. 



\section{Algorithms and regret bounds for binary distributions}
\label{sec:algo}

In this section we present algorithms and regret bounds for the \kmax problem with max value-index feedback with arm outcomes according to binary distributions. We first show some properties of reward functions which are crucial for our regret analysis. We then present an algorithm for the case when the ordering of $v_1,\ldots,v_n$ values is known. 
In this case, we will see that the problem can be reduced to a CMAB-T instance solvable using the standard CUCB method. 
Then we consider the case when the ordering of $v_1,\ldots, v_n$ is a priori unknown. We present an algorithm and show that this algorithm achieves the same regret bound as when the ordering is known up to constant factors.

For the convenience of exposition, we assume that values $v_1, \ldots, v_n$ are distinct. This ensures that for any action $S_t$, there is a unique arm achieving the maximum value over the arms in $S_t$. This is equivalent to allowing for non-unique values and using a deterministic tie-breaking rule. More discussion on this can be found in Appendix~\ref{equ:equiv}.


We define an extended set $\mathcal{B}$ of arms as the union of two sets of arms, namely $\mathcal{Z}$ and $\mathcal{V}$. The first set $\mathcal{Z} = \{a_1^z,\ldots,a_n^z\}$ consists of $n$ arms whose outcomes are according to independent Bernoulli random variables $Z_1,\ldots, Z_n$ with mean values $p_1,\ldots,p_n$. The second  $\mathcal{V} = \{a_1^v,\ldots, a_n^v\}$ consists of $n$ arms whose outcomes are deterministic with corresponding values $v_1,\ldots,v_n$. Note that the outcome of each arm $i\in \arms$ can be written as $X_i = V_i Z_i$. Each time an action $S_t$ is played, we obtain information for some of arms in $\mathcal{B}$. We call these arms \emph{triggered arms}, and observe their outcome values as feedback. An important observation is that when arms are ordered in decreasing order with respect to their values, lower-order arms being triggered implies that higher-order arms had zero outcomes. We define $T^z_{i,t}$ as the number of triggering times for arm $a_i^z$ and $T^v_{i,t}$ as the number of triggering times for arm $a_i^v$ up to time step $t$. 

\paragraph{Notation} For any two vectors $\bm{x},\bm{x}'\in \mathbb{R}^n$, we write $\bm{x}\geq \bm{x}'$ if $x_i\geq x'_i$ for all $i\in [n]$.

\subsection{Properties of reward functions} \label{sec:prop}

For the prevailing case of binary distributions of arm outcomes, for any set $S\in \mathcal{F}$, under assumption that arms in $\arms$ are ordered in decreasing order of values $v_1,\ldots, v_n$, the expected reward can be expressed as
\begin{equation} 
\label{eqn:exp-reward}
r_S(\bm{p},\bm{v}) = \sum_{i\in S} v_i p_i \prod_{j\in S: j<i}(1-p_j),
\end{equation}
where any product over empty set has value $1$.

There are two key properties of the reward function in (\ref{eqn:exp-reward}) that we leverage in our regret analysis, namely monotonicity and smoothness, which we define and show to hold in the following. 

\paragraph{Monotonicity} The first property is monotonicity. 
\begin{lemma} \label{lem:mono}
For every $S\in \mathcal{F}$, $r_S(\bm{p},\bm{v})$ is increasing in every $p_i$ and $v_i$.
\end{lemma}
It is clear from  (\ref{eqn:exp-reward}) that $r_S(\bm{p},\bm{v})$ is monotonic increasing in $v_i$. It can be shown that it is monotonic increasing in $p_i$ by taking first derivative with respect to $p_i$ and showing that it is  non negative.



\paragraph{Smoothness} The second property is \emph{relative triggering probability modulated} (RTPM) smoothness, which with a slight abuse of notation is defined as follows, for an arbitrary set of arms $\mathcal{B}$ with expected outcome values $\bm{\mu}$, and $r_S(\bm{\mu})$ denoting the expected reward of action $S$.






\begin{definition}[RTPM smoothness] \label{defn:rTPM}
The RTPM smoothness holds for reward functions $r_S$ if, for any two vectors of expected outcomes $\bm{\mu}$ and $\bm{\mu}'$, for every action set $S$,
$$|r_S(\bm{\mu}) - r_S(\bm{\mu}')|\leq \sum_{i\in S}q_i^{\bm{\mu},S}b_i|\mu_i - \mu_i'|,$$
where $b_i$ is a constant per-arm weight and 
$q_i^{\bm{\mu},S}$ denotes the triggering probability of arm $i$ under action $S$ and the expected arm outcomes $\bm{\mu}$. Note that when $r_S(\bm{\mu})$ is increasing in $\bm{\mu}$, and $\bm{\mu}\geq \bm{\mu}'$, we can remove the absolute value in the above inequality.
\end{definition}






The definition of RTPM smoothness is a generalization of the condition in \cite{Wang2017} extended to have the triggering probability $q_i^{\bm{\mu},S}$ and weight $b_i$ to modulate the standard 1-norm condition. 
This allows us to account for arm-specific values as we will see in our next lemma. The intuition is that we underweight the importance of arms with small triggering probability or weight in the expected reward. Even if for some arm $i$ we cannot estimate its expected value accurately, we lose very little in the expected reward. This will be an important point in our regret analysis.





Let $q_i^{\bm{p},S}$ be the triggering probability for arm $a_i^z$ and action $S$ and $\tilde{q}_i^{\bm{p},S}$ be the triggering probability for arm $a_i^v$ and action $S$. 
Note that $q_i^{\bm{p},S} = (1-p_1)(1-p_2)\cdots(1-p_{i-1})$
and $\tilde{q}_i^{\bm{p},S} = (1-p_1)(1-p_2)\cdots(1-p_{i-1})p_i$. Note that arm $a_i^z$ is triggered when the winner arm has value smaller than or equal to $v_i$, 
while arm $a_i^v$ is triggered when arm $i\in \arms$ is the winner arm, thus $\tilde{q}_i^{\bm{p},S} = q_i^{\bm{p},S} p_i$.


The following is a key lemma for our regret analysis. 


\begin{lemma} \label{lem:general-case} 
The expected reward functions in (\ref{eqn:exp-reward}) satisfy the RTPM condition with respect to the extended set $\mathcal{B}$ of arms, i.e. for every $S$, $\bm{p}$ and $\bm{p}'$, $\bm{v}$, and $\bm{v'}$
, it holds 
$$
|r_S(\bm{p},\bm{v}) - r_S(\bm{p}',\bm{v}')|
		\leq  2\sum_{i\in S}q_i^{\bm{p},S}v_i'|p_i-p_i'|  + \sum_{i\in S}\tilde{q}_i^{\bm{p},S}|v_i-v_i'|.
$$

Furthermore, if $\bm{p}\geq \bm{p}'$, then we can remove the factor 2 in the last inequality. 
\end{lemma}



The lemma can be intuitively explained as follows. When an arm $i\in \arms$ has small value $v_i$ and the corresponding arm $a_i^z$ is unlikely to be triggered (small $q$), its importance in regret analysis diminishes. On the other hand, if the arm is unlikely to win (small $\tilde{q}$), it is also not important in our analysis. This concept is important for the proof of our regret bounds. For arms with small values or arms whose values are unlikely to be observed, we may not be able to estimate their value and probability parameters accurately. The lemma suggests  this is not a critical issue for regret analysis.


To prove Lemma \ref{lem:general-case}, we consider a sequence of vectors changing from $(\bm{p},\bm{v})$ to $(\bm{p}',\bm{v}')$ and add up the changes in expected rewards. The full proof is provided in Appendix~\ref{app:TPMcondition}.

\subsection{Algorithm for known ordering of values}

We propose the CUCB algorithm defined in Algorithm~\ref{alg:simple}. The estimates of parameters $\bm{p}$ and $\bm{v}$ are initialized to vectors with all elements equal to $1$. At each time step in which $v_j$ is observed to be the maximum value in the selected action, we update the estimate of $v_j$ and the estimates for $p_i$, for arms in the action set ordered before $j$.  The algorithm maintains an upper confidence bound (UCB) for both parameters and feeds the UCB values to the approximation oracle to determine the next action. 
We note that for this case, our problem can be interpreted as a conjunctive cascading bandit~\cite{Kveton2015a} with binary-valued arms. The ordering of arms within each action enables us to observe values of all arms ordered before the winner, which makes the problem easier to solve than when the ordering of values is a priori unknown to the algorithm.



\begin{algorithm}[t] 
\caption{CUCB algorithm for known ordering of values}
\begin{algorithmic}[1]
	\State For $i\in \arms$, $T_i \gets 0$ \Comment{Number of triggering times for $a_i^z$}
	\State For $i\in \arms$, $\hat{p}_i\gets 1$, $\hat{v}_i\gets 1$ \Comment{Initial estimation of parameters}
	\For{$t=1,2,\ldots,T$}
	\State For $i\in \arms$, $\rho_i \gets \sqrt{\frac{3\log(t)}{2T_i}}$ \Comment{Confidence radius values}
	\State For $i\in \arms$, $\bar{p}_i \gets \min\{\hat{p}_i+\rho_i, 1\}$, $\bar{v}_i\gets \hat{v}_i$	\Comment{UCB values}
	\State $S\gets \mathrm{Oracle}(\bm{\bar{p}},\bm{\bar{v}})$	\Comment{
		Oracle decides the next action}
	\State Play $S$ and observe winner index $i^*$ and value $v_{i^*}$
	\State Update $\hat{v}_{i^*}$ for winner arm $i^*$: $\hat{v}_{i^*} \gets v_{i^*}$  
	\State For $i\in S$ such that $i\leq i^*$: $T_i \gets T_i+1$
	\State For $i\in S$ such that $i<i^*$: $\hat{p}_i \gets (1-1/T_i)\hat{p}_i$
	\State $\hat{p}_{i^*} \gets (1-1/T_{i^*})\hat{p}_{i^*} + 1/T_{i^*}$
	\EndFor
\end{algorithmic}
\label{alg:simple}
\end{algorithm}


For each action $S\in \mathcal{F}$, we define the gap $\Delta_S = \max\{\alpha\mathrm{OPT}(\bm{p},\bm{v}) - r_S(\bm{p},\bm{v}),0\}$. We call an action $S$ \textit{bad} if $\Delta_S > 0$. For each arm $i\in \arms$ that is contained in at least one bad action, we define
$$
\Delta_{\min}^i = \min_{S:i\in S, q_i^{\bm{p},S},\tilde{q}_i^{\bm{p},S}>0, \Delta_S > 0}\Delta_S \ \hbox{ and } \ \Delta_{\max}^i = \max_{S:i \in S,
q_i^{\bm{p},S},\tilde{q}_i^{\bm{p},S}>0, \Delta_S > 0}\Delta_S.
$$
For every arm $i\in \arms$ that is not contained in a bad action, we define $\Delta_{\min}^i  = \infty$ and $\Delta_{\max}^i =0$. Let $\Delta_{\min} = \min_{i\in[n]}\Delta_{\min}^i$ and $\Delta_{\max} = \max_{i\in[n]}\Delta_{\max}^i $.


\begin{theorem} \label{thm:bound-simple}
For the \kmax problem with max value-index feedback, under assumption that ordering of values is a priori known to the algorithm and $\Delta_{\min}>0$, Algorithm~\ref{alg:simple} has the following distribution-dependent regret bound,
$$
	R(T)\leq  c_1\sum_{i=1}^n\left(\frac{k}{\Delta_{\min}^i} + \log\left(\frac{k}{\Delta_{\min}^i}+1\right)\right)\log(T)
	 +c_2\sum_{i=1}^n \left(\left(\log\left(\frac{ k}{\Delta_{\min}^i}+1\right)+1
	\right)\Delta_{\max}+ 1\right),
$$
where $c_1$ and $c_2$ are some positive constants. 
\end{theorem}

The regret upper bound in Theorem~\ref{thm:bound-simple} implies the regret upper bound $O((nk/\Delta_{\min})\log(T))$ which is comparable with the regret upper bound for the standard CMAB-T problem \cite{Chen2016}. This in turn is tight with respect to dependence on $T$ in comparison with the lower bound in \cite{Kveton2015}. In Theorem~\ref{thm:bound-simple}, the only term in the regret bound that depends on horizon time $T$ is the first summation term. In this summation term, the summands have two terms, one scaling linearly with $k/\Delta_{\min}^i$ and other scaling logarithmically with $k/\Delta_{\min}^i$, which are due to uncertainty of parameters $\bm{p}$ and $\bm{v}$, respectively. Hence, we may argue that the uncertainty about values of parameters $\bm{p}$ has more effect on regret than uncertainty about values of parameters $\bm{v}$. The regret bound in Theorem~\ref{thm:bound-simple} implies a $\tilde{O}(\sqrt{T})$ distribution-independent regret bound. 



To see how the regret analysis of the algorithm can be decomposed to two CMAB-T problems, we consider the contribution of each action to regret, i.e, $\Delta_{S_t} = \max\{\alpha\ \mathrm{OPT}(\bm{p},\bm{v}) - r_{S_t}(\bm{p},\bm{v}),0\}$. Let $\mathcal{F}_t$ be the good event $\{r_{S_t}(\bar{\bm{p}}, \bar{\bm{v}}) \geq \alpha\ \mathrm{OPT}(\bm{p},\bm{v}) \}$ meaning that the approximation oracle works well. 
By the smoothness condition, under $\mathcal{F}_t$ we have 
$$
\Delta_{S_t} \leq r_{S_t}(\bar{\bm{p}}_t,\bar{\bm{v}}_t) - r_{S_t}(\bm{p},\bm{v}) \leq \sum_{i\in S_t}q_i^{\bm{p},S}\bar{v}_{i,t}(\bar{p}_{i,t} - p_i) + \sum_{i\in S_t}\tilde{q}_i^{\bm{p},S}(\bar{v}_{i,t} - v_i).
$$

Clearly, the first term corresponds to regret from the set of arms $\mathcal{Z}$, and the second term corresponds to regret from the set of arms $\mathcal{V}$. We bound $\Delta_{S_t}$ by bounding the two summation terms individually. The first summation term is standard in existing literature. For bounding the second term, we need to make extra steps as our estimates for $v_i$ are not more and more accurate as the number of selections of arm $i$ increases. The UCB value for $v_i$ remains at the upper bound value $1$ until arm $a_i^v$ is triggered once and we then know the exact value of $v_i$. We show the full proof in Appendix~\ref{pr:bound-simple}.


\subsection{Algorithm for unknown ordering of values}

We consider the case when the agent does not know the ordering of values $\{v_i \mid i\in S_t\}$ for every action $S_t$. This greatly reduces the information that can be deduced from observed information feedback. To see this, we consider the stage of arm $a_i^v$ before its value $v_i$ is observed. 
Note that in the simpler case when the ordering of the values is known, $a_i^z$ is triggered whenever arm $i\in \arms$ is ordered before the winner arm $j \in \arms$. This is because in this case we can deduce that $Z_i$ has to be zero, since otherwise arm $i$ with a higher value $v_i$ would beat arm $j$ and $j$ cannot be the winner. 
However, since the ordering is unknown in the general case, we can no longer carry out the above deduction and it is unclear whether $Z_i$ has value $0$ or $1$.
More specifically, suppose that in round $t$ we play action $S_t$, and arm $j \in S_t$ with value $v_j$ is the winner and value-index pair $(v_j,j)$ is observed. For an arm $i\in S_t$, we have not observed $v_i$ so do not know whether $v_i < v_j$ or $v_i > v_j$. For the first case, arm $i$ could take a non-zero value that is not observed, while it takes zero value for the other case. Importantly, we note that the triggering of $a_i^z$ is dependent on whether knowing the value of $v_i$ or not. This is different from the CMAB framework and thus we cannot simply reduce this setting back to an equivalent CMAB setting.


A naive approach is to adopt the CUCB algorithm for the simpler case and introduce $T_i^v$ as the triggering time for arm $a_i^v$. We update parameters of arm $i\in \arms$ only when $T_i^v \neq 0$. However, this approach could fail for each arm $i$ with large $v_i$ and small $p_i$. Note that the estimate of $p_i$ will not be updated until $v_i$ is observed. However, the upper bound of $1$ for $p_i$ is clearly an overestimate for this type of arms, which would cause large regrets during the period when their values are not observed. This will be reflected as an undesirable factor in the regret upper bound.

We propose a variant of the CUCB algorithm in Algorithm~\ref{algo:final}. In this algorithm, we do not wait to update $p_i$ only after observing value $v_i$. 
We start with optimistic initial estimates $\hat{v}_i=1$, which means we treat $a_i^z$ as always triggered at the beginning and every arm has a high chance of being a winner. 
In this way, the true winners will gradually stand out, while we are still giving chances to those arms whose values have not been observed yet. This intuitively makes sense as even if $v_i$ takes value $1$, the above-mentioned type of arms will not be important for our regret analysis due to their small probability parameter estimates.


\begin{algorithm}[t] 
\caption{CUCB algorithm for unknown ordering of values}
\begin{algorithmic}[1]
	\State For $i\in \arms$, $T_i \gets 0$, $\tilde{T}_i \gets 0$ \Comment{Number of triggering times for $a_i^z$ and $a_i^v$}
	\State For $i\in \arms$, $\hat{p}_i\gets 1$, $\hat{v}_i\gets 1$ \Comment{Initial estimation of parameters}
	\For{$t=1,2,\ldots,T$}
	\State For $i\in \arms$, $\rho_i \gets \sqrt{\frac{3\log(t)}{2T_i}}$, $\tilde{\rho}_i \gets \mathbf{1}\{\tilde{T}_i = 0\}$ 
	\State For $i\in \arms$, $\bar{p}_i \gets\min\{\hat{p}_i+\rho_i, 1\}$, $\bar{v}_i\gets \min\{\hat{v}_i+\tilde{\rho}_i, 1\}$	\Comment{UCB values}
	\State $S\gets \mathrm{Oracle}(\bm{\bar{p}},\bm{\bar{v}})$	\Comment{
		Oracle decides the next action}	
	\State Play $S$ and observe winner index $i^*$ and value $v_{i^*}$
	\If{$\tilde{T}_{i^*}=0$} 
	\State \text{Reset } $T_{i^*}
	\gets 0, \quad \tilde{T}_{i^*}\gets  1
	, \quad\hat{v}_{i^*} \gets v_{i^*}$
	\EndIf
	\State For $i\in S$ such that $\hat{v}_i\geq v_{i^*}$: $T_i \gets T_i+1$
	\State For $i\in S$ such that $\hat{v}_i> v_{i^*}$: $\hat{p}_i \gets (1-1/T_i)\hat{p}_i$
	\State For $i\in S$ such that $\hat{v}_i= v_{i^*}$: $\hat{p}_i \gets (1-1/T_i)\hat{p}_i+1/T_i$
	\EndFor
\end{algorithmic}
\label{algo:final}
\end{algorithm} 


We have the following regret upper bound for the case of unknown ordering of values.


\begin{theorem} \label{thm:bound-general}
For the \kmax problem with max value-index feedback, under assumption that ordering of values is unknown to the algorithm a priori and $\Delta_{\min}>0$, Algorithm~\ref{algo:final} has the following distribution-dependent regret bound,
$$
R(T)\leq c_1 \sum_{i=1}^n\left(\frac{k}{\Delta_{\min}^i} + \log\left(\frac{k}{\Delta_{\min}^i}+1\right)\right) \log(T) +c_2 \sum_{i=1}^n \left(\left(\log \left(\frac{ k}{\Delta_{\min}^i}+1\right)+1\right)\Delta_{\max}+1\right),
$$

for some positive constants $c_1$ and $c_2$.



\end{theorem}



The regret upper bound in Theorem~\ref{thm:bound-general} implies regret upper bound $O((nk/\Delta_{\min})\log(T))$, which agrees with the bound for the simpler case in Theorem~\ref{thm:bound-simple} up to constant factors. The regret bound in Theorem~\ref{thm:bound-general} implies a $\tilde{O}(\sqrt{T})$ distribution-independent regret bound. 




\paragraph{Proof sketch} In the following we give a sketch of the proof of Theorem \ref{thm:bound-general}. The full proof is provided in Appendix \ref{pr:bound-general}.

Our problem does not fit into the standard CMAB-T framework. As discussed above, the algorithm assumes that $a_i^z$ is triggered and takes value zero. This may not be the ground truth in the case when $v_i$ is actually less than the winner value. 
Therefore, the estimates are biased. 
We cannot simply apply the regret result of CMAB-T or follow its analysis to reach our result. To tackle this difficulty, we introduce the concept of \textit{arm equivalence}. In each round $t$, for every arm $i$ with parameters $(p_i,v_i)$ and $T_{i,t}^v = 0$, we replace it with an equivalent arm with parameters $(p'_i,v'_i)$ where $p'_i = p_iv_i$ and $v'_i = 1$. This equivalence is in the sense that two arms have equal expected outcome values. 

We use similar framework for regret analysis as for the CUCB algorithm. However, note that one of the key assumptions fails to hold in our setting, i.e., we do not always have upper confidence bounds for parameters $p_i$. Thus, we developed new technical steps to account for this. 

Firstly, we notice the following fact about item equivalence.

\begin{lemma} \label{lem:equi}
For every set $S\in \mathcal{F}$, 
$r_S(\bm{p},\bm{v})\leq r_S(\bm{p}',\bm{v}')$.
\end{lemma}

Then we consider the contribution of action $S_t$ to regret, i.e. $\Delta_{S_t}$, under the good event $\mathcal{F}_t$ that the approximation oracle works well, i.e. $r_{S_t}(\bar{\bm{p}}, \bar{\bm{v}})\geq \alpha\ \mathrm{OPT}(\bar{\bm{p}}, \bar{\bm{v}})$.
By Lemma \ref{lem:equi}, for each $t$ such that $1\leq t\leq T$ we have,
\begin{equation} \label{ineq:opt}
\alpha\ \mathrm{OPT}(\bm{p}'_t, \bm{v}'_t)\geq \alpha\ \mathrm{OPT
}(\bm{p},\bm{v}).
\end{equation}
Thus,
\begin{align*}
\Delta_{S_t} &\leq \alpha\ \mathrm{OPT}(\bm{p}'_t, \bm{v}'_t) - r_{S_t}(\bm{p},\bm{v})\\
&\leq \alpha\ \mathrm{OPT}(\bm{p}'_t, \bm{v}'_t) - r_{S_t}(\bm{p},\bm{v}) + r_{S_t}(\bm{\bar{p}}, \bm{\bar{v}})- \alpha\ \mathrm{OPT}(\bar{\bm{p}}, \bar{\bm{v}})\\
&\leq r_{S_t}(\bar{\bm{p}}, \bar{\bm{v}}) -  r_{S_t}(\bm{p},\bm{v})\\ 
&= (r_{S_t}(\bar{\bm{p}}, \bar{\bm{v}}) - r_{S_t}(\bm{p}'_t,\bm{v}'_t) ) + (r_{S_t}(\bm{p}'_t, \bm{v}'_t) - r_{S_t}(\bm{p},\bm{v}) ),
\end{align*}
where the first inequality is due to condition~(\ref{ineq:opt}), the second inequality is due to the approximation oracle, and the third inequality is due to monotonicity of $r_S$ in $\bm{p}$ and $\bm{v}$. We call the term inside first bracket as the regret caused by estimation error $\Delta^{e}_{S_t}$, 
and the term inside the second bracket as the regret caused by replacement error $\Delta^{r}_{S_t}$. To obtain a tight regret upper bound, we require that the regret caused by replacement error over the time horizon $T$ is not greater than the that by estimation error, i.e, under a series of good events,
$
\sum_{t=1}^{T}\Delta^{r}_{S_t}\leq\sum_{t=1}^{T}\Delta^{e}_{S_t} 
$.
This would justify the intuition of using replacement arms. Now we look closely at these two terms separately. 

By Lemma~\ref{lem:general-case}, we have 
\begin{equation*} \label{eqn:est-err}
	\Delta^{e}_{S_t}\leq \sum_{i\in S_t}q_i^{\bm{p}',S}\bar{v}_{i,t}(\bar{p}_{i,t} - p'_{i,t}).
\end{equation*}
Note that we do not need to include the $v_i$ term as $v'_{i,t} = \bar{v}_i =1$ for all $i$ when $v_i$ is not observed, and $v'_{i,t} = \bar{v}_i = v_i$ after $v_i$ is observed. In both cases, there is no estimation error for $v_i$. 

We also apply Lemma~\ref{lem:general-case} to the second summation term to obtain
\begin{equation*} \label{eqn:rep-err}
\Delta^{r}_{S_t}\leq 2\sum_{i\in S_t}q_i^{\bm{p},S}v'_{i,t}(p_i - p'_{i,t})+ \sum_{i\in S_t}\tilde{q}_i^{\bm{p},S}(v'_{i,t}-v_i).
\end{equation*}
To sum up, we have

$$
\Delta_{S_t}  \leq  \sum_{i\in S_t}q_i^{\bm{p}',S}\bar{v}_{i,t}(\bar{p}_{i,t} - p'_{i,t})  + 2\sum_{i\in S_t}q_i^{\bm{p},S}v'_{i,t}(p_i - p'_{i,t}) + \sum_{i\in S_t}\tilde{q}_i^{\bm{p},S}(v'_{i,t}-v_i).
$$



We can bound the first term by following the proof of the regret bound for the standard CMAB-T problem, stated in Theorem \ref{thm:standard-bound} for completeness. To see this, recall that we have reset the counts $T_i$ and the estimates $p_i$ at the time observing $v_i$. This is because $p'_{i,t} = p_iv_i$ when $v_i$ is unknown and $p'_{i,t} = p_i$ afterwards. However, for both stages, our estimates are accurate such that $p_i'$ always lies within the confidence interval which decreases as the counter number increases.

For the second term, we note that $p'_{i,t} = p_iv_i$ in the first stage, and $p'_i = p_i$ after observing $v_i$. Therefore, the contribution to regret by the second term is zero in the second stage. For the first stage, this term can be analyzed in a similar way as the last term. The key observation is that $p_i - p'_{i,t} = p_i(1-v_i)\leq p_i$. This will be the key to removing the extra factor.

Finally, we note that the analysis for the last summation term is the same as the case where the ordering of values is known, since there is no change to the triggering process of arms in $\mathcal{V}$.

Summing up the resulting bounds over time horizon $T$, we can prove our theorem.



\section{Arbitrary distributions of arm outcomes with finite supports}
\label{sec:ext}
\begin{algorithm}[t]
	\caption{CUCB algorithm for arbitrary distributions with finite supports}
	\begin{algorithmic}[1]
		\State For $i\in \arms$, $\sigma(i)\gets 0$ \Comment{Number of known values for arm $i$}
		\State For $i\in \arms$, $T_{i,0} \gets 0$, $\tilde{T}_{i,0} \gets 0$ \Comment{Number of triggering times for the fictitious arm}
		
		\State For $i\in \arms$, $\hat{p}_{i,0}\gets 1$, $\hat{v}_{i,0}\gets 1$ \Comment{Initial estimates of parameters for the fictitious arm}
		\For{$t=1,2,\ldots$}
		\State For $i\in \arms$ and $j\in [\sigma(i)]$, $\rho_{i,j} \gets \sqrt{\frac{3\log(t)}{2T_{i,j}}}$, $\tilde{\rho}_{i,j} \gets \mathbf{1}\{\tilde{T}_{i,j} = 0\}$ \Comment{Confidence radius of parameters}
		\State For $i\in \arms$ and $j\in [\sigma(i)]$, $\bar{p}_{i,j} \gets\min\{\hat{p}_{i,j}+\rho_{i,j}, 1\}$, $\bar{v}_{i,j}\gets \min\{\hat{v}_{i,j}+\tilde{\rho}_{i,j}, 1\}$\Comment{UCB values of parameters}
		\State Transform $\bar{\bm{p}}$ to $\tilde{\bm{p}}$ using Eq. (\ref{eqn:trans})
		\State $S\gets \mathrm{Oracle}(\tilde{\bm{p}},\bar{\bm{v}})$	\Comment{
			Oracle $k$-MAX PTAS decides the next action}	
		\State Play $S$ and observe winner index $i^*$ and value $v$
		\If{$v\notin\{\hat{v}_{i^*,j},j\in [\sigma(i^*)]\}$} 
		\State $\sigma(i^*)\gets \sigma(i^*)+1, \quad T_{i^*,\sigma(i^*)}\gets 0, \quad \tilde{T}_{i^*,\sigma(i^*)}\gets  1
		, \quad\hat{v}_{i^*,\sigma(i^*)} \gets v$
		\EndIf
		\State For $i\in S$ and $j\in [\sigma(i)]$ such that $\hat{v}_{i, j}\geq v$: $T_{i, j} \gets T_{i, j}+1$
		\State For $i\in S$ and $j\in [\sigma(i)]$ such that $\hat{v}_{i, j}> v$: $\hat{p}_{i, j} \gets (1-1/T_{i, j})\hat{p}_{i, j}$
		\State For $i\in S$ and $j\in [\sigma(i)]$ such that $\hat{v}_{i, j}= v$: $\hat{p}_{i, j} \gets (1-1/T_{i, j})\hat{p}_i+1/T_{i, j}$
		\EndFor
	\end{algorithmic}
	\label{algo:ext-general}
\end{algorithm} 
We consider the more general case of arbitrary discrete distributions of arm outcomes with finite supports, which accommodates the case of binary distributions as a special case. We show that it is possible to represent a variable corresponding to an arm outcome with a set of binary variables. This allows us to extend to the general case of discrete distributions with finite supports.

To see this, let $X_i$ be a random variable with an arbitrary discrete distribution with finite support as defined in Section~\ref{sec:form}. Recall that $\Pr[X_i = v_{i,j}] = p_{i,j}$, for $j\in \{0,1,\ldots,s_i\}$, where $v_{i,j}\in [0,1]$ and $v_{i,0}=0$. Let $X_{i,1}, \ldots, X_{i,s_i}$ be independent binary random variables such that $X_{i,j}$ takes value $v_{i,j}$ with probability $\tilde{p}_{i,j}$, and value $0$ otherwise, with
\begin{equation} \label{eqn:trans}
\tilde{p}_{i,j} = \left\{
\begin{array}{ll}
	\frac{p_{i,j}}{1-\sum_{l=j+1}^{s_i} p_{i,l}} & \hbox{if }1\leq j < s_i\\
	p_{i,s_i} & \hbox{if } j=s_i.
\end{array}
\right .
\end{equation}

It can be readily checked that $\max\{X_{i,j} \mid j\in [s_i]\}$ has the same distribution as the original random variable $X_i$. In this way, we establish the equivalence between binary variables and any discrete variables with a finite support in terms of the $\max$ operator. This means that we can use our algorithm to solve the $k$-MAX bandit problem for any discrete distributions of arm outcomes with finite supports.

For example, if we know all the possible values, we can order them, and then use the algorithm with known value orders. Otherwise, we can use the algorithm with unknown value orders. We present our Algorithm \ref{algo:ext-general} for the general discrete distributions with finite support. The algorithm is an extension of Algorithm \ref{algo:final}, with slight modifications that allow us to relax the assumption of knowing the support sizes of arm distributions.



Recall that we work with binary arms with outcomes according to $\{X_{i,j}, i\in [n], j\in [s_i]\}$ where $s_i+1$ is the support size of $X_i$. The key is that we introduce a counter $\sigma(i)$ to denote the number of observed values of $X_i$ and dynamically maintain a list of values for $X_i$. 
We increase this counter and reset the triggering times and probability estimates for the $\sigma(i)$-th arm whenever we observe a new value for $X_i$. On the other hand, we use a fictitious arm with value 1 as placeholder for those arms whose values remain unobserved. Since we have no information on the support size, we always keep this fictitious arm and update its probability estimates whenever arm $i$ is selected in an action.   

Note that we convert UCBs of the binary arms to multi-valued forms according to relationship in Equation~(\ref{eqn:trans}) and use the $k$-MAX PTAS in \cite{Chen2016} as the offline oracle. We give further explanations justifying this usage as follows. In the equivalent binary form, we would need an oracle such that for each binary arm $(i,j)$ with outcome $X_{i,j}$, if $(i,j)$ is selected, then all $(i,j')$ for $j'\in[s_i]$ must also be selected. Because of the fact that $X_i = \max\{X_{i,j}\mid j\in [s_i]\}$, we just need to convert $\bar{\bm{p}}$ to $\bar{\bm{q}}$ and use the $k$-MAX PTAS as the offline oracle for the equivalent binary case.

\section{Numerical results} \label{sec:numeric}

\begin{figure*}[t]
\begin{center}
	\includegraphics[width=0.29\linewidth]{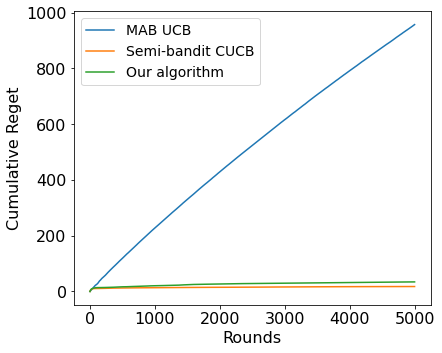}\hspace*{5mm}
	\includegraphics[width=0.29\linewidth]{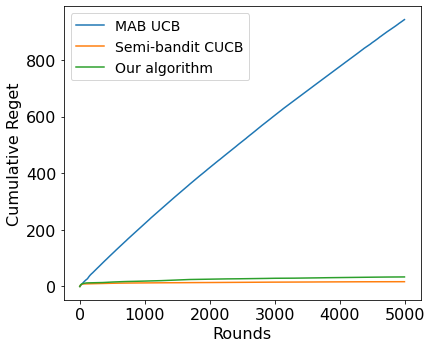}\hspace*{5mm}
	\includegraphics[width=0.29\linewidth]{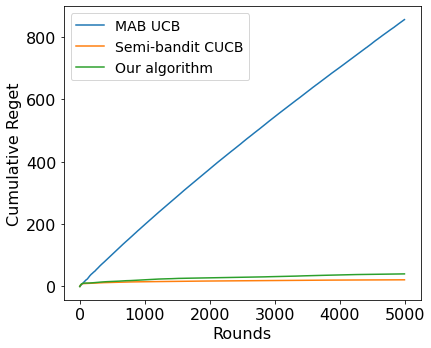}\vspace*{-2mm}
\end{center}
\caption{Cumulative regret versus the number of rounds for Algorithm~\ref{algo:final} for different distributions of arm outcomes defined in Section~\ref{sec:numeric}.}
\label{fig:reg-test}
\end{figure*}

We perform experiments to evaluate performance of Algorithm \ref{algo:final} on some specific problem instances. We compare our algorithm with two baseline methods: the well-known UCB algorithm treating each set of size $k$ as one arm, and standard CUCB algorithm for semi-bandit CMAB setting. We use the greedy algorithm as the offline oracle. The code we use is available on GitHub: \url{https://github.com/Sketch-EXP/kmax}.

\paragraph{Setup} We consider settings with $n=9$ arms and sets of cardinality $k=3$. We tested on three different distributions of arm outcomes representing different scenarios.
\begin{enumerate}	
	\item[{\bf D1}] $\bm{v} = (0.1, 0.2, \ldots, 0.9)^\top$. $\bm{p}$ is such that $p_i=0.3$, for $1\leq i\leq 6$, and $p_i = 0.5$, otherwise. 
	\item[{\bf D2}] Compared to D1, we introduce an arm $i$ with small $v_i$ and large $p_i$. Specifically, for arm $1$ we redefine $p_1= 0.9$ and keep $v_1$ unchanged. 
	\item[{\bf D3}] Compared to D1, we introduce an arm $i$ with large $v_i$ and small $p_i$. Specifically, for the last arm we redefine $p_9 = 0.2$ and keep $v_9$ unchanged. Note that arm 9 has the same expected value as arm 6, but arm 9 is in the optimal set.
\end{enumerate}

Note that the optimal action is $S^* = \{7,8,9\}$ in all cases. Distributions D1, D2 and D3 represent different scenarios. D1 is the base case. In D2, there is a stable arm with low value, while in D3 there is a high-risk high-reward arm. Both are not easy to observe and cause challenges for our algorithm design, especially the latter type of arms, which can outperform less-risky arms under the maximum value reward function. 

We expect our algorithm to perform well for all three cases. 
We run each experiment for horizon time $T=5,000$. In each round, we select arms according to the offline oracle and sample their values for updates. We compare the reward to that of the optimal action $S^*$. We repeat the experiments for $20$ times and compute average cumulative regrets.

We show the cumulative regrets of Algorithm \ref{algo:final} and two baseline methods in Figure \ref{fig:reg-test}. We plot the 1-approximation regrets instead of $(1-1/e)$-approximation regret as the offline greedy oracle performs much better than $(1-1/e)$-approximation in this case. 
We can see that our algorithm achieves much lower regret compared to the UCB algorithm, that is close to that of the CUCB method under semi-bandit CMAB setting. This confirms that comparable regret can be achieved to that under more informative semi-bandit feedback.



\section{Conclusion}\label{sec:conc}

We studied the \kmax combinatorial multi-armed bandit problem under a new max value-index feedback. 
We proposed a CUCB algorithm for the case when the ordering of values is a priori known, and for the other case, we proposed a new algorithm based on the concept of arm equivalence. We showed that our algorithms guarantee a regret upper bound that is matching that under more informative semi-bandit feedback up to constant factors. 

Future work may consider whether the same regret bound can be achieved for the \kmax problem under full-bandit feedback. 
It would also be of interest to consider other CMAB problems under feedback structures that lie in between semi-bandit and full-bandit feedback.


%
\bibliography{ref}

\appendix
\onecolumn



\section{CMAB-T framework and additional notation} \label{standard-cmab}


We review the framework and results for the classical CMAB problem with triggered arms considered by \cite{Wang2017}. In this problem, the expected reward is a function of action $S$ and expected values of arm outcomes $\bm{\mu}$. Denote the probability that action $S$ triggers arm $i$ as $p_i^{\bm{\mu},S}$. It is assumed that in each round the values of triggered arms are observed by the agent. The CUCB algorithm~\cite{Chen2013} is used to estimate the expectation vector $\bm{\mu}$ directly from samples. 

The following theorem is for the standard CMAB problem with triggered arms (CMAB-T). It is assumed that the CMAB-T problem instance satisfies monotonicity and 1-norm TPM bounded smoothness (Definition \ref{defn:rTPM} with $b_i = B$ for all $i\in [n]$). We will use some proof steps and the result of this theorem in proofs of our results. 




\begin{theorem}[Theorem 1 in \cite{Wang2017}] \label{thm:standard-bound}
	For the CUCB algorithm on a CMAB-T problem instance satisfying monotonicity and 1-norm TPM bounded smoothness, we have the following distribution-dependent regret bound,
	$$R(T)\leq 576B^2k\left(\sum_{i=1}^n \frac{1}{\Delta_{\min}^i}\right)\log(T) +\left(\sum_{i=1}^n \left(\log\left(\frac{2Bk}{\Delta_{\min}^i}+1\right)+2\right)\right)\frac{\pi^2}{6}\Delta_{\max}+4Bn$$
	where $\Delta_{\min} = \inf_{S:i\in S, p_i^{\bm{\mu},S}>0, \Delta_S > 0}\Delta_S$.
\end{theorem}


\label{sec:events}


We next give various definitions used in our analysis. The definitions are given specifically for binary distributions of arm outcomes.

We define two sets of \textit{triggering probability (TP) groups}. Let $j$ be a positive integer. For the set $\mathcal{Z}$ of arms we define the triggering probability group $\mathcal{S}_{i,j}^z$ as
$$
\mathcal{S}_{i,j}^z = \{S\in \mathcal{F} \mid 2^{-j}<q_i^{\bm{p},S}\leq 2^{-j+1}\}.
$$
We define the triggering probability group $\mathcal{S}_{i,j}^v$ for the set $\mathcal{V}$ of arms similarly. We note that the triggering probability groups divide actions that trigger arm $i$ into separated groups such that the actions in the same group contribute similarly to the regret bound. 

Let $N_{i,j,t}^z$ be the counter of the cumulative number of times $i$ in TP group $S_{i,j}^z$ is selected at the end of round $t$. Similarly, we use $N_{i,j,t}^v$ to denote the counter of the cumulative number of times $i$ in TP group $S_{i,j}^v$ is selected at the end of round $t$. 

We define the \textit{event-filtered regret} for a sequence of events $\{\mathcal{E}_t\}$ as
$$
R(T,\{\mathcal{E}_t\}) = T\ \alpha\ \mathrm{OPT}(\bm{p},\bm{v}) - \E\left[\sum_{t=1}^{T} \mathbf{1}(\mathcal{E}_t)r_{S_t}(\bm{p},\bm{v})\right]
$$
which means that the regret is accounted for in round $t$ only if event $\mathcal{E}_t$ occurs in round $t$.

We next define four good events as follows:
\begin{itemize}
	\item[{\bf E1}] The approximation oracle works well, i.e.
	$$
	\mathcal{F}_t = \{r_{S_t}(\bar{\bm{p}}, \bar{\bm{v}}) \geq \alpha\ \mathrm{OPT}(\bar{\bm{p}}, \bar{\bm{v}})\}.
	$$
	\item[{\bf E2}] The parameter vector $\bm{p}$ is estimated well, i.e. for every $i\in [n]$ and $t\geq 1$,
	$$
	\mathcal{N}_{i,t} = \{|\hat{p}_{i,t-1} - p_i| <\rho_{i,t}\} 
	$$
    where $\hat{p}_{i,t-1}$ is the estimator of $p_i$ at round $t$ and $\rho_{i,t} := \sqrt{3\log(t)/(2T_{i,{t-1}})}$. We denote this event with $\mathcal{N}_t$. 
	\item[{\bf E3}] Triggering is nice for $\mathcal{Z}$ given a set of integers $\{j^i_{\max}\}_{i\in [n]}$, i.e. 
	for every TP group $\mathcal{S}_{i,j}^z$ defined by arm $i$ and $1\leq j\leq j_{\max}^i$, under the condition $\sqrt{6\log(t)/(N_{i,j,t-1}^z/3) 2^{-j} }\leq 1$, it holds 
	$T_{i,t-1}^z\geq \frac{1}{3}N_{i,j,t-1}^z 2^{-j}$.
	We denote this event with $\mathcal{G}^z_t$. 
	
	\item[{\bf E4}] Triggering is nice for $\mathcal{V}$, i.e. for every arm $i\in \arms$, under the condition $N_{i,j,t-1}^z\geq 3p_i^{-1}\log(t) 2^j$,
	there is $T^v_{i,t-1}\neq 0$. Equivalently, we can define this event in terms of TP group $S^v_{i,j}$ by removing the factor of $p_i^{-1}$, i.e., under the condition $N_{i,j,t-1}^v\geq 3\log(t) 2^j$. We denote this event with ${\mathcal{G}}^v_t$. 
\end{itemize}

Note that E1, E2 and E3 are also used in \cite{Wang2017}, and event E4 is newly defined for our problem. We can easily show the following bound by using the Hoeffding's inequality. 
\begin{lemma}
	For each round $t\geq 1$, it holds $\Pr(\neg \mathcal{G}^v_t )\leq n/t^2$.
\end{lemma}



\section{Proofs}



\subsection{Equivalence between the deterministic tie-breaking rule and the unique value assumption}
\label{equ:equiv}
In Section~\ref{sec:algo}, we assume that all values $v_1, v_2, \ldots, v_n$ are unique, and state that this is equivalent to allowing for non-unique values with a deterministic tie-breaking rule.
In this appendix, we provide a more formal argument for this equivalence.

Deterministic tie-breaking means that whenever we have two arms $i$ and $j$ appearing in the same action $S$ and their outcomes have the same values $v_i=v_j$ and this outcome is the maximum outcome
	among all outcomes of arms in $S$, we deterministically decide one of $i$ or $j$ would win according to a predefined order.
Without loss of generality, we assume the predefined order is that the smaller indexed arm always wins in tie-breaking, that is, if $v_i=v_j$, then arm $i$ wins if $i < j$.

%

For any problem instance with the above tie-breaking rule, we convert it to another instance where all values are unique, and show that the two instances essentially behave the same.
The conversion is essentially adding lower order digits to the current value of $v_i$'s to make all values unique while respecting the deterministic tie-breaking rule.
The actual conversion is as follows.
Suppose a problem instance in the deterministic tie-breaking setting is $(\bm{p},\bm{v})$, and without loss of generality, assume that $v_1 \ge v_2 \ge \cdots \ge v_n$.
Let $\varepsilon_0 > 0$ be an arbitrarily small value.
Let $\varepsilon = \varepsilon_0 / n$.
Then we construct a new problem instance $(\bm{p}',\bm{v}')$, such that $\bm{p}'=\bm{p}$, and for all $i$, $v'_i = v_i + (n-i) \varepsilon$. 
With this construction, we know that if $v_i > v_j$, then $i<j$, and $v'_i = v_i + (n-i) \varepsilon > v_j + (n-j)\varepsilon = v'_j$; 
	if $v_i = v_j$ and $i < j$, $v'_i = v_i + (n-i) \varepsilon > v_j + (n-j)\varepsilon = v'_j$.
This means we have all unique values with $v'_1 > v'_2 > \cdots > v'_n$, and the value order respects the previous value order together with the deterministic tie-breaking rule.
Because $\bm{p}'=\bm{p}$, we know that for any action $S$, the winning index distribution for these two instances must be the same.
Moreover, for every $i$, $|v'_i - v_i| = (n-i) \varepsilon < \varepsilon_0$, which means the reward value of the new instance is arbitrarily close to the original instance.
Therefore, through the above construction, we show that any problem instance under the deterministic tie-breaking rule can be viewed as an instance with unique values that
	are arbitrarily close to the original values.
This shows that the problem instances under the deterministic tie-breaking rule can be treated equivalently as instances with unique values with essentially the same result.

\subsection{Proof of Lemma \ref{lem:mono}}


Without loss of generality, consider $S = \{1,\ldots, k\}$ under assumption $v_1\geq \cdots \geq v_k$. The expected regret can be written as 
$$
r_S(\bm{p}, \bm{v}) = p_1v_1+(1-p_1)p_2v_2 +\cdots+(1-p_1)\cdots(1-p_{k-1})p_kv_k.
$$
It is clear from the expression that $r_S(\bm{p},\bm{v})$ is monotonic increasing in $v_i$.

Let us consider $r_S(\bm{p},\bm{v})$ for arbitrarily fixed $\bm{v}$. Taking derivative with respect to $p_i$, we have
$$
\frac{d}{dp_i} r_S(\bm{p},\bm{v}) = (1-p_1)\cdots(1-p_{i-1})\left[v_i - p_{i+1}v_{i+1}- \left(\sum_{j>i}(1-p_{i+1})\cdots (1-p_{j})p_{j+1}v_{j+1}\right)\right].
$$

We claim that the term inside the bracket is non negative. This can be shown as follows,
\begin{eqnarray*} 
	& & v_i - p_{i+1}v_{i+1} - \left(\sum_{j>i}(1-p_{i+1})\cdots (1-p_{j})p_{j+1}v_{j+1}\right)\\ 
	& \geq& v_i(1-p_{i+1}) - (1-p_{i+1})p_{i+2}v_{i+2}-(1-p_{i+1})(1-p_{i+2})p_{i+3}v_{i+3}-\cdots\\
	& \geq &(1-p_{i+1})(1-p_{i+2})\cdots(1-p_{k-1})(v_i - p_kv_k)\\ 
	& \geq &(1-p_{i+1})(1-p_{i+2})\cdots(1-p_{k})v_i\\
	& \geq & 0
\end{eqnarray*}
Thus, the reward function is monotonic increasing in $p_i$.

\subsection{Proof of Lemma \ref{lem:general-case}}
\label{app:TPMcondition}



For the purpose of this proof only, we assume that the arms are ordered in decreasing order of values. This applies to both $\bm{v}$ and $\bm{v}'$. Recall that
$$r_S(\bm{p},\bm{v}) = \sum_{i\in S} v_i p_i \prod_{j\in S: j<i}(1-p_j)$$

Let $\bm{p} = (p_1,\ldots,p_k)$ and  $\bm{p'} = (p_1',\ldots,p_k')$, and $\bm{p}^{(0)} = \bm{p}'$, $\bm{p}^{(j)} = (p_1,\ldots, p_j, p_{j+1}',\ldots, p_k')$, for $1\leq j < k$, and $\bm{p}^{(k)} = \bm{p}$. Similarly, let $\bm{v} = (v_1,\ldots,v_k)$ and  $\bm{v}' = (v_1',\ldots,v_k')$, and $\bm{v}^{(0)} = \bm{v}'$, $\bm{v}^{(j)} = (v_1,\ldots, v_j, v_{j+1}',\ldots, v_k')$, for $1\leq j < k$, and $\bm{v}^{(k)} = \bm{v}$. 

Let us define
$$f(\bm{p}^{(j)}, \bm{v}^{(j)}) = p_1v_1 + \cdots+(1-p_1)\cdots(1-p_{j-1})p_jv_j+ (1-p_1)\cdots(1-p_{j})p_{j+1}'v_{j+1}'+ \cdots
$$
Clearly, we have $r_S(\bm{p},\bm{v}) = f(\bm{p}^{(k)}, \bm{v}^{(k)})$ and $r_S(\bm{p}',\bm{v}') = f(\bm{p}^{(0)}, \bm{v}^{(0)})$.

Note that 
$$
f(\bm{p}^{(j-1)}, \bm{v}^{(j-1)}) = p_1v_1 + \cdots+(1-p_1)\cdots(1-p_{j-1})p_j'v_j'+ (1-p_1)\cdots(1-p_{j}')p_{j+1}'v_{j+1}'+ \cdots
$$

The only difference is caused by position $j$. By definition of triggering probabilities $q_i^{\bm{p}, S}$ and $\tilde{q}_i^{\bm{p},S}$ we can write,
\begin{align*}
	|f(\bm{p}^{(j)}, \bm{v}^{(j)})-f(\bm{p}^{(j-1)}, \bm{v}^{(j-1)})| 	
	=&|q_j^{\bm{p},S}(p_jv_j - p_j'v_j' - \sum_{i>j}(1-p_{j+1}')\ldots(1-p_{i-1}')p_{i}'v_i'(p_j -p_j')|\\
	\leq& q_j^{\bm{p},S}p_j|v_j - v_j'|+ q_j^{\bm{p},S}v_j'|p_j - p_j'| \\
	&+ q_j^{\bm{p},S}(p_{j+1}'v_{j+1}'+(1-p_{j+1}')p_{j+2}'v_{j+2}'+\ldots)|p_j - p_j'|\\
	\leq& 2q_j^{\bm{p},S}v_j'|p_j-p_j'|  + \tilde{q}_j^{\bm{p},S}|v_j-v_j'|
\end{align*}
where the first inequality is due to the triangle inequality and the second inequality is due to the monotonicity property. If $\bm{p}\geq \bm{p}'$, we only need to consider the first two terms in the second line. 

Now we note that
\begin{align*}
|r_S(\bm{p},\bm{v}) - r_S(\bm{p}',\bm{v}')|&= |f(\bm{p}^{(k)}, \bm{v}^{(k)})-f(\bm{p}^{(0)}, \bm{v}^{(0)})| \\
&\leq\sum_{j=1}^{k} |f(\bm{p}^{(j)}, \bm{v}^{(j)})-f(\bm{p}^{(j-1)}, \bm{v}^{(j-1)})| 
\end{align*}

Summing up over $j$ we can obtain the statement of the lemma.


\subsection{Proof of Theorem \ref{thm:bound-simple}}
\label{pr:bound-simple}

We consider the contribution of action $S_t$ to regret $\Delta_{S_t}$, for every $t\geq 1$. Let $M_i := \Delta_{\min}^i$. Recall that $\Delta_S = \max\{\alpha\cdot\mathrm{OPT}_{\bm{p},\bm{v}} - r_S(\bm{p},\bm{v}),0\}$. Assume that $\Delta_{S_t}\geq M_{S_t}$ where $M_{S} = \max_{i\in S}M_i$. Note that if $\Delta_{S_t}<M_{S_t}$ then $\Delta_{S_t}=0$, since we have either an empty set, or $\Delta_{S_t}<M_{S_t}<M_i$ for some $i\in S_t$.

By the smoothness condition, we have
$$
\Delta_{S_t}\leq r_{S_t}(\bar{\bm{p}}_t,\bar{\bm{v}}_t) - r_{S_t}(\bm{p},\bm{v})\leq \sum_{i\in S_t}q_i^{\bm{p},S}\bar{v}_{i,t}(\bar{p}_{i,t} - p_i) + \sum_{i\in S_t}\tilde{q}_i^{\bm{p},S}(\bar{v}_{i,t} - v_i).
$$
Since $\Delta_{S_t}\geq M_{S_t}$, we add and subtract $M_{S_t}$ from the last expression and we have,
\begin{align*}
	\Delta_{S_t}&\leq-M_{S_t}+2\left(\sum_{i\in S_t}q_i^{\bm{p},S}\bar{v}_{i,t}(\bar{p}_{i,t} - p_i)+ \sum_{i\in S_t}\tilde{q}_i^{\bm{p},S}(\bar{v}_{i,t} - v_i)\right)\\
	&\leq 2\left(\sum_{i\in S_t}q_i^{\bm{p},S}\bar{v}_{i,t}(\bar{p}_{i,t} - p_i) - \frac{M_i}{4k}\right) + 2\left(\sum_{i\in S_t}\tilde{q}_i^{\bm{p},S}(\bar{v}_{i,t} - v_i) - \frac{M_i}{4k}\right).
\end{align*}
Let us call the first term $\Delta_{S_t}^p$ and the second term $\Delta_{S_t}^v$. We bound $\Delta_{S_t}$ by bounding the two summation terms individually.

Note that we can bound the term $\Delta_{S_t}^p$ following the same procedure as in the proof for Theorem~\ref{thm:standard-bound}. However, we cannot use the same procedure for $\Delta_{S_t}^v$. The key difference is that our estimate for $v_i$ will not be more and more accurate as the number of selections of arm $i$ increases. We know the exact value of $v_i$ as soon as it is triggered once.  We assume that arm $i$ is in TP group $S_{i,j}^v$. Let $j_i$ be the index of the TP group with $S_t\in S_{i,j_i}^v$. We take $j^i_{\max} = \log(4k/M_i+1)$.
\begin{itemize}
	\item Case 1: $1\leq j_i\leq j^i_{\max}$. In this case, $\tilde{q}_i^{\bm{p},S}\leq 2\cdot 2^{-j_i}$,
	$$\tilde{q}_i^{\bm{p},S}(\bar{v}_{i,t} - v_i)\leq 2\cdot 2^{-j_i}\ \mathbb{1}\{T_{i,t}^v=0\}$$
	Under the good event $\mathcal{G}^v_t$, when $N_{i,j_i,t-1}^v\geq 3\log(t)\cdot 2^j$, the contribution to regret is zero.  Otherwise, it is bounded by,
	$$\tilde{q}_i^{\bm{p},S}(\bar{v}_{i,t} - v_i)\leq 2\cdot 2^{-j_i}.$$
	
	\item Case 2: $j_i\geq j^i_{\max}+1$. In this case,
	$$\tilde{q}_i^{\bm{p},S}(\bar{v}_{i,t} - v_i)\leq 2\cdot 2^{-j_i}\leq \frac{M_i}{4k}.$$
	Thus the term does not contribute to regret in this case.
\end{itemize}

We next calculate the filtered regret under the good events mentioned above and the event that $\Delta_{S_t}\geq M_{S_t}$. Note that 
$$
R(T,\{\Delta_{S_t}\geq M_{S_t}\},\mathcal{F}_t, \mathcal{N}_t, \mathcal{G}^z_t, \mathcal{G}^v_t)\leq \sum_{t=1}^{T}\Delta_{S_t}^p + \sum_{t=1}^{T}\Delta_{S_t}^v.
$$

For the event E3 we set $j^i_{\max} = \log(4 k/M_i+1)$. By  Theorem~\ref{thm:standard-bound} in Appendix \ref{standard-cmab}, we know that under good events E1, E2 and E3, the first term is bounded by
$$
\sum_{t=1}^{T}\Delta_{S_t}^p \leq 1152k \left(\sum_{i=1}^k \frac{1}{M_i}\right)\log(T) + 4n
$$

and, otherwise, the corresponding filtered regrets are 
\begin{align*} \label{eqn:good-event}
	&R(T,\neg\mathcal{F}_t)\leq (1-\beta)T\cdot\Delta_{max} \\
	&R(T,\neg\mathcal{N}_t)\leq \frac{\pi^2}{3}n\Delta_{\max} \\
	&R(T, \neg\mathcal{G}^z_t)\leq  \frac{\pi^2}{6}\left(\sum_{i=1}^{n} \log\left(\frac{4k}{M_i}+1\right)\right)\Delta_{\max}.
\end{align*}
Now we focus on bounding the contribution of the second term to regret. Note that under event E4,
$$\sum_{t=1}^{T}\Delta_{S_t}^v  = \sum_{i=1}^{n}\sum_{j=1}^{\infty}\sum_{s=0}^{N_{i,j,T-1}^v}\kappa_{j_i,T}(M_i, s)$$
where $$
\kappa_{j,T}(M, s) = \left\{
\begin{array}{ll}
	2\cdot 2^{-j} & \hbox{ if }s < 3\log(T)\cdot 2^j\\
	0 & \hbox{ if } s \geq 3\log(T)\cdot 2^j
\end{array}
\right . .
$$
For every arm $i$ and $j\geq 1$, we have
\begin{align*}
	\sum_{s=0}^{N_{i,j,T-1}^v}\kappa_{j_i,T}(M_i, s)\leq \sum_{s=0}^{3\log (T)\cdot 2^{j_i}}\kappa_{j_i,T}(M_i, s) = 6\log(T).
\end{align*}
Hence, the contribution of the second term to regret is bounded by 
$$
\sum_{i=1}^{n}\sum_{j=1}^{\infty}\sum_{s=0}^{N_{i,j,T-1}^v}\kappa_{j_i,T}(M_i, s)\leq 6\left(\sum_{i=1}^{n}  \log\left(\frac{4k}{M_i}+1\right)\right)\log(T).
$$
The filtered regrets for the case when event E4 fails to hold is bounded by,
$$
R(T,\neg \mathcal{G}^v_t)\leq \sum_{i=1}^{T}\Pr(\neg \mathcal{G}^v_t )\Delta_{max}\leq \frac{\pi^2}{6}n\Delta_{\max}.
$$
We obtain the distribution-dependent regret bound by adding up the filtered regrets calculated above. The corresponding distribution-independent regret bound is implied by taking $M_i = \sqrt{16nk/T}$ for every $i\in [n]$.

\subsection{Proof of Lemma \ref{lem:equi}}
Without loss of generality assume that $S=[k]$ and $v_1\geq v_2\geq\cdots\geq v_k$. Recall that we can write $$
r_S(\bm{p}, \bm{v}) = p_1v_1+(1-p_1)p_2v_2 +\ldots+(1-p_1)\ldots(1-p_{k-1})p_kv_k.
$$
Now for $\bm{p} = (p_1,\ldots,p_k)$ and  $\bm{p}' = (p_1',\ldots,p_k')$, let 
$$\bm{p}^{(j)} = (p'_1,\ldots,p'_j,p_{j+1},\ldots,p_k)$$
and define similarly $\bm{v}^{(j)}$ for $\bm{v}$ and $\bm{v}'$.
After changing $p_1$ to $p'_1 = p_1v_1$ and $v_1$ to $v'_1 = 1$, 
$$r_S(\bm{p}^{(1)}, \bm{v}^{(1)}) = p_1v_1+(1-p_1v_1)p_2v_2 +\cdots+(1-p_1v_1)\cdots(1-p_{k-1})p_kv_k.$$
Clearly we have $r_S(\bm{p}^{(1)}, \bm{v}^{(1)})\geq r_S(\bm{p},\bm{v})$. Following the same argument, we can see that $r_S(\bm{p}^{(2)}, \bm{v}^{(2)})\geq r_S(\bm{p}^{(1)}, \bm{v}^{(1)})$. Continuing this way to $r_S(\bm{p}^{(k)}, \bm{v}^{(k)})$ we can prove the lemma.

\subsection{Proof of Theorem \ref{thm:bound-general}}
\label{pr:bound-general}
By the RTPM smoothness condition, as discussed in the main text, we have
\begin{equation*}
	\Delta_{S_t}\leq \sum_{i\in S_t}q_i^{\bm{p}',S}\bar{v}_{i,t}(\bar{p}_{i,t} - p'_{i,t})  + 2\sum_{i\in S_t}q_i^{\bm{p},S}v'_{i,t}(p_i - p'_i)+ \sum_{i\in S_t}\tilde{q}_i^{\bm{p},S}(v'_{i,t}-v_i).
\end{equation*}
\paragraph{Key step: bounding the contribution of each action to regret}
Let $M_i = \Delta_{\min}^i$. Assume that $\Delta_{S_t}\geq M_{S_t}$ where $M_{S} = \max_{i\in S}M_i$. As in the known value ordering case, we can bound $\Delta_{S_t}$ such that,
\begin{align} \label{reg-parts}
	\Delta_{S_t}&\leq-M_{S_t}+2\left(\sum_{i\in S_t}q_i^{\bm{p}',S}\bar{v}_{i,t}(\bar{p}_{i,t} - p'_{i,t})+ 2\sum_{i\in S_t}q_i^{\bm{p},S}v'_{i,t}(p_i - p'_i) + \sum_{i\in S_t}\tilde{q}_i^{\bm{p},S}(v'_{i,t}-v_i)\right)\nonumber \\ 
	&\leq 2\left[\left(\sum_{i\in S_t}q_i^{\bm{p}',S}\bar{v}_{i,t}(\bar{p}_{i,t} - p'_{i,t}) - \frac{M_i}{8k}\right) + 2\left(\sum_{i\in S_t}q_i^{\bm{p},S}v'_{i,t}(p_i - p'_i) -\frac{M_i}{8k} \right)+ \left(\sum_{i\in S_t}\tilde{q}_i^{\bm{p},S}(v'_{i,t} - v_i) - \frac{M_i}{4k}\right)\right].
\end{align}

Let $j_i$ be the index of the TP group of $S_t$ such that $S_t\in S_{i,j_i}^z$. We bound $\Delta_{S_t}$ by bounding the three summation terms in (\ref{reg-parts}) separately.
\begin{itemize}
	\item 
	Bounding the first term. 
	Recall that we reset $T_{i,t}^z$ and $N^z_{i,j,t}$ at the time we observe $v_i$. This is because $p'_{i,t} = p_iv_i$ and $v'_{i,t} = 1$ when $v_i$ is unknown (first stage) and $p'_{i,t} = p_i$ and $v'_{i,t} = v_i$ after observing $v_i$ (second stage). A key observation is that within both stages our estimates are accurate in the sense that under event E2, the approximation error decreases as the counter number increases in the following way.
	$$\bar{p}_{i,t} - p'_{i,t}\leq 2\rho_i = 2\sqrt{\frac{3\log(t)}{2T^z_{i,t-1}}}.$$
	We note that for the second stage where $v_i$ is observed, this term is similar to the $\Delta^p_{S_t}$ term in the known value ordering case. Specifically, under event E3,
	$$\bar{p}_{i,t} - p'_{i,t}\leq \min\left\{\sqrt{\frac{18\log(t)}{2^{-j_i}\cdot N^z_{i,j_i,t-1}}},1\right\}$$
	and
	$$q_i^{\bm{p}', S}\bar{v}_{i,t}(\bar{p}_{i,t} - p'_{i,t})\leq \min\left\{\sqrt{\frac{72\cdot 2^{-j_i}\log(T)}{N^z_{i,j_i,t-1}}},2\cdot 2^{-j_i}v_i\right\}.$$
	For the event E3, let $j^i_{\max} = \log(8 k/M_i+1)$. In the case $j_i\geq j^i_{\max}+1$, this term does not contribute to regret as we have,
	$$q_i^{\bm{p}', S}(\bar{p}_{i,t} - p'_{i,t}) \leq 2\cdot 2^{-j_i}\leq \frac{M_i}{8k}$$
	Similarly, for $1\leq j_i\leq j^i_{\max}$, there is no contribution to regret if $N^z_{i,j,t-1}\geq l_{j_i,T}(M_i)$ where
	$$l_{j,T}(M)= \left\lfloor\frac{4608\cdot 2^{-j}k^2\log(T)}{M^2}\right\rfloor.$$
	For the first stage, we note that the event E3 is not required to hold, as we treat $a^z_i$ as always triggered and the triggering is always nice for $\mathcal{Z}$ in the first stage. Thus for the first stage we have, 
	$$q_i^{\bm{p}', S}\bar{v}_{i,t}(\bar{p}_{i,t} - p'_{i,t})\leq \min\left\{\sqrt{\frac{6\log(T)}{T^z_{i,t-1}}},1\right\}$$
	This term does not contribute to regret if $T^z_{i,t-1}\geq l'_{T}(M_i)$ where
	$$l'_{T}(M)= \left\lfloor\frac{384 k^2\log(T)}{M^2}\right\rfloor. $$
	
	
	\item 
	Bounding the second term.
	Take $j^i_{\max} = \log(8k/M_i+1)$. Similarly as the first term, in the case $j_i\geq j^i_{\max}+1$, the contribution to regret is non-positive. For $1\leq j_i\leq j^i_{\max}$, as $p'_{i,t} = p_iv_i$, we have
	$$q_i^{\bm{p}, S}v'_{i,t}(p_i - p'_{i,t})\leq 2\cdot2^{-j_i}p_i(1-v_i).$$
	Under the event E4, we know that the contribution to regret is zero if $N^z_{i,j_i,t-1}\geq 3p_i^{-1}\log(T)\cdot 2^j$. Otherwise, it is upper bounded by $2\cdot2^{-j_i}p_i$.
	
	\item 
	Bounding the third term. Take $j^i_{\max} = \log(8k/M_i+1)$. In the case $j_i\geq j^i_{\max}+1$, the contribution to regret is non-positive. For $1\leq j_i\leq j^i_{\max}$, we have $\tilde{q}_i^{\bm{p},S}\leq 2\cdot 2^{-j_i}p_i$, thus
	$$\tilde{q}_i^{\bm{p},S}(\bar{v}_{i,t} - v_i)\leq 2\cdot2^{-j_i}p_i\tilde{\rho}_i= 2\cdot2^{-j_i}p_i\cdot\mathbb{1}\{T^v_{i,t}=0\}.$$
	Under the event E4, we know that the contribution to regret is zero if $N^z_{i,j_i,t-1}\geq 3p_i^{-1}\log(T)\cdot 2^j$. Otherwise, it is upper bounded by
	$$\tilde{q}_i^{\bm{p},S}(\bar{v}_{i,t} - v_i)\leq 2\cdot 2^{-j_i}p_i$$
	Note that this bound is the same as the bound for the second term.
\end{itemize}

\paragraph{Summing over the time horizon}
Next, we sum up $\Delta_{S_t}$ over time $T$ and calculate the filtered regret under the above mentioned good events and the event that $\Delta_{S_t}\geq M_{S_t}$, i.e, $R(\{\Delta_{S_t}\geq M_{S_t}\},\mathcal{F}_t, \mathcal{N}_t, \mathcal{G}^z_t, \mathcal{G}^v_t)$. 

By equation (\ref{reg-parts}), we know that the filtered regret can be upper bounded by sum of three terms over the time horizon $T$. By Theorem~\ref{thm:standard-bound} in Appendix \ref{standard-cmab}, we know under good events E1, E2 and E3, the second stage of the first term is bounded by
$$
2\left(\sum_{i\in S_t}q_i^{\bm{p},S}\bar{v}_{i,t}(\bar{p}_{i,t} - p'_{i,t}) - \frac{M_i}{8k}\right) \leq
2304k\left(\sum_{i=1}^n\frac{1}{M_i}\right)\log(T) + 4n
$$
and, otherwise, the corresponding filtered regrets are bounded by, 
\begin{align*} 
	&R(T,\neg\mathcal{F}_t)\leq (1-\beta)T\Delta_{\max} \\
	&R(T,\neg\mathcal{N}_t)\leq \frac{\pi^2}{3}n\Delta_{\max} \\
	&R(T, \neg\mathcal{G}^z_t)\leq  \frac{\pi^2}{6}\sum_{i=1}^{n} \log\left(\frac{8k}{M_i}+1\right)\Delta_{\max}.
\end{align*}
To bound the first term, we also need to derive a bound for the first stage when the value is not observed. This is bounded by,
	$$2\left(\sum_{i\in S_t}q_i^{\bm{p}',S}(\bar{p}_{i,t} - p'_{i,t}) - \frac{M_i}{8k}\right) \leq \sum_{i=1}^n \sum_{s=1}^{l'_{T}(M_i)} 2\sqrt{\frac{6\log(T)}{s}}
	=64 k \left(\sum_{i=1}^n \frac{1}{M_i}\right)\log(T).$$


Following the similar analysis as for the $\Delta^v_{S_t}$ term of Theorem \ref{thm:bound-simple}, we know that

$$2\sum_{t=1}^{T}\left(\sum_{i\in S_t}\tilde{q}_i^{\bm{p},S}(v'_{i,t}-v_i) - \frac{M_i}{4k}\right)\leq 12\sum_{i=1}^n \log\left(\frac{8k}{M_i}+1\right) \log(T).$$


Similarly, we can bound
$$
4\sum_{t=1}^{T}\left(\sum_{i\in S_t}q_i^{\bm{p},S}v'_{i,t}(p_i - p'_{i,t}) -\frac{M_i}{8k} \right)\leq 24\sum_{i=1}^{n}\log\left(\frac{8k}{M_i}+1\right) \log(T).
$$

The filtered regret for the case where event E4 fails to hold is bounded by,
$$
R(T,\neg \mathcal{G}^v_t)\leq \sum_{i=1}^{T}\Pr(\neg \mathcal{G}^v_t )\Delta_{\max}\leq \frac{\pi^2}{6}n\Delta_{\max}.
$$

We obtain the distribution-dependent regret by adding up the filtered regrets calculated above. Similarly as before, the distribution-dependent regret is implied by taking $M_i = \sqrt{64nk/T}$ for every $i\in [n]$.

\end{document}